\documentclass{article}
\usepackage{acra}

\usepackage{graphicx}
\usepackage{amsmath,amssymb} 
\usepackage{color}
\usepackage{array}
\usepackage{outlines}
\usepackage{caption} 
\usepackage{algorithmic}
\usepackage{algorithm}
\usepackage{hyperref}
\usepackage{cleveref}
\usepackage{wrapfig}
\usepackage{subfigure}
\usepackage{booktabs}
\usepackage{balance}
\usepackage{tabularx}

\begin{document}

\title{Deep Similarity Metric Learning for Real-Time Pedestrian Tracking}

\author{Michael Thoreau and Navinda Kottege}

\maketitle

\begin{abstract}
Tracking by detection is a common approach to solving the Multiple Object Tracking problem. In this paper we show how learning a deep similarity metric can improve three key aspects of pedestrian tracking on a multiple object tracking benchmark. We train a convolutional neural network to learn an embedding function in a Siamese configuration on a large person re-identification dataset. The offline-trained embedding network is integrated in to the tracking formulation to improve performance while retaining real-time performance. The proposed tracker stores appearance metrics while detections are strong, using this appearance information to: prevent ID switches, associate tracklets through occlusion, and propose new detections where detector confidence is low. This method achieves competitive results in evaluation, especially among online, real-time approaches. We present an ablative study showing the impact of each of the three uses of our deep appearance metric.

\end{abstract}

\section{Introduction}
Accurately tracking objects of interest such as pedestrians and vehicles in video streams is an important problem with applications in many fields such as surveillance, robotics and autonomous vehicles. The problem of Multiple Object Tracking (MOT) in video has mostly been addressed in recent literature using the `tracking by detection' framework. In this formulation, detections are combined to estimate the trajectories of tracked objects. Solutions can generally be grouped in to online and batch processes. The difference being, online solutions use measurements only as they arrive while a batch process may build globally optimal trajectories by considering measurements at all times.

In this paper we present an online approach to solving the MOT problem for pedestrian tracking and evaluate it on the MOTChallenge dataset \cite{Leal-TaixeMOTChallenge2015Benchmark2015,MilanMOT16BenchmarkMultiObject2016}. 

Motivated by the large amounts of labelled data now available for pedestrian re-identification problems, the proposed method uses a deep-learning approach to appearance modelling. We present a convolutional neural network, trained in a Siamese configuration to produce a discriminative appearance similarity metric for pedestrians.

We present three ways in which this deep appearance metric learning can be used in MOT and show how using two of these components together can achieve competitive performance on a tracking benchmark. We compare our results to those of other methods and evaluate each use of the proposed appearance metric independently in an ablative study. First we show how a learned appearance metric can be used to improve the \textit{assignment} of candidate detections to form short tracks (tracklets) as the first step in creating longer optimal tracks. Next we show how the same metric learner can perform \textit{detection boosting} to reduce false negatives where detections are missing within a person's track. Lastly the deep appearance metric is used to perform iterative appearance based merging of tracklets to form longer tracks, a process we call \textit{tracklet association}. We accomplish this as an online process, with a playback delay of only a few seconds, at a frame rate suitable for real-time applications.

The rest of the paper is organised as follows; section~\ref{sec:relatedwork} describes the related approaches in the literature, section~\ref{sec:SMDT} introduces the proposed Siamese Deep Metric Tracker, section~\ref{sec:evaluation} evaluates the proposed method on the publicly available MOT16 dataset of the MOTChallenge, section~\ref{sec:discussion} discusses the evaluation results and section~\ref{sec:conclusions} concludes the paper.

\section{Related Work}
\label{sec:relatedwork}
Solutions to the multiple object tracking problem fall in to two distinct categories; batch and online processing. In batch processing,  detections are combined in a global sense, rather than frame by frame, to form optimal tracks\cite{RenFasterRCNNRealTime2015,LiuSSDSingleShot2016,Tychsen-SmithDeNetScalableRealtime2017}. Despite the apparent performance advantages of batch methods as described by Luo et al. in their extensive literature review\cite{LuoMultipleObjectTracking2014a}, we consider only online methods in this work, where filtered tracks are available with little to no delay, motivated by potential real-time applications in surveillance, robotics and industrial safety.

In some online approaches, tracklet states are estimated by a probabalistic model such as a Kalman filter \cite{BochinskiHighSpeedtrackingbydetectionusing2017,BewleySimpleOnlineRealtime2016}. Others have used deep learning to learn to estimate the motion of tracked objects from data, including estimating the birth and death of tracks\cite{MilanOnlineMultiTargetTracking2016}.

A difficult aspect of tracking by detection is solving the data association problem present when grouping detections or merging tracklets. Some works is present in the literature that use confidence estimation to aid in the data association problem by prioritising high confidence tracks\cite{BaeConfidenceBasedDataAssociation2017,BaeRobustOnlineMultiobject2014}. Others leverage image information, where even simple appearance modelling has been shown to make data association more robust\cite{takala2007multi}. Appearance modelling plays a larger role in single object tracking, where only appearance is used to track objects given a prior\cite{BolmeVisualobjecttracking2010}. 

More recently, the availability of labelled data has motivated methods utilising deep learning for appearance modelling. Siamese networks have been used for single object tracking to great effect by Feichtenhofer et al. and Liu et al., where the deep appearance model is used to search successive frames\cite{FeichtenhoferDetectTrackTrack2017,TaoSiameseInstanceSearch2016}. 

In multiple object tracking, online learning has been used to discriminate between tracked objects based on appearance albeit at limited speed due to computational complexity\cite{BaeConfidenceBasedDataAssociation2017}.
 
Some methods using deep learning have achieved outstanding results on the MOT Challenge \cite{LiuEndtoEndComparativeAttention2017,YuPOIMultipleObject2016}. For example, \cite{Leal-TaixeLearningtrackingSiamese2016} achieve good results with a siamese network, however with a requirement for a gallery of images to be stored from past tracks to do re-identification \cite{Leal-TaixeLearningtrackingSiamese2016}. Wojke et al.  solves this by using a deep similarity metric learning network to store a gallery of metrics \cite{WojkeSimpleOnlineRealtime2017}. He et al. goes one step further and uses a deep recurrent network to compute an appearance metric which incorporates temporal information from the tracked object, to good effect \cite{HeSOTMOT2017}.

Informed by the literature, we have developed our `Siamese Deep Metric Tracker'; with competitive tracking accuracy and real-time performance, crucial for applications in robotics.


    \begin{figure}[b!]
        \centering
        \includegraphics[width=0.8\columnwidth]{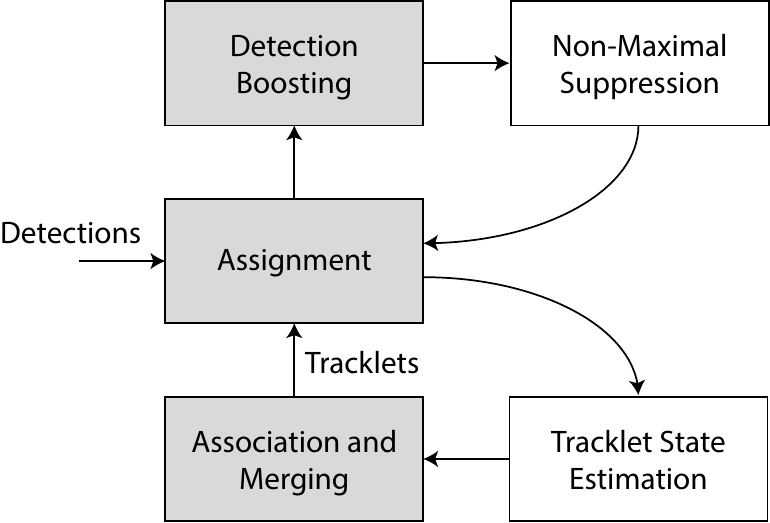}
        \caption{The proposed process where \textit{Assignment}, \textit{boosting}, and \textit{tracklet association} components benefit from the use of deep appearance modelling.}
        \label{fig:OSTAB}
    \end{figure}

\section{Siamese Deep Metric Tracker}
\label{sec:SMDT}
Here we present our proposed Siamese Deep Metric Tracker to perform online multiple object tracking. A strong appearance model is central to this proposed method. We use a single deep neural network, detailed in section~\ref{sec:deep_metric}, to enable or assist three components of our object tracking algorithm shown at a high level in figure \ref{fig:OSTAB}. We solve the problem in multiple stages; firstly, detections are \textit{Assigned} to tracklets, as detailed in section~\ref{sec:assignment}; detections are then \textit{Boosted} as described in section~\ref{sec:boosting}; and finally tracklets are \textit{Associated} as described in section~\ref{sec:Tracklet_Association}.

    

    \subsection{Notation}
    We use the following notation in all equations, explanations and algorithm listings in this paper. Let the set of estimated tracklets be $\mathcal{T}$, containing $J$ tracklets $T_j$. Let the estimated state of tracklet $j$ at time $t$ be $T_j^t$ and the predicted state of tracklet be ${T_j^t}^{\prime}$. Let a set of detections at time $t$ be $\mathcal{D}^t$ containing $I$ detections $D_i$.

    \subsection{Deep Similarity Metric}
    \label{sec:deep_metric}
    A robust appearance model can improve simple object tracking by preventing tracks from drifting to false positive detections, and by enabling objects to be tracked through occlusion.
    
    At each time step we compute a feature vector $f \in \mathbb{R}^{128}$ for each candidate detection in a single batch. Computing all features in a batch is an efficient use of GPU resources, taking only $\approx 20\text{\,ms}$ for a typical batch of 40 image patches.
    The network, with layers listed in table \ref{tab:nn_structure}, uses pre-trained convolutional layers from VGG-16\cite{SimonyanVeryDeepConvolutional2014}, followed by two fully connected layers with batch and $l_2$ normalisation on the output layer. The use of pre-trained networks as feature extractors in Siamese/triplet networks has been shown to reduce the number of iterations required for convergence and improve accuracy\cite{HermansDefenseTripletLoss2017}. Euclidean distance between feature vectors lying within a unit hypersphere measures the distance $d_a = || f_1-f_2 ||$ between two input patches in the appearance similarity space. The appearance affinity $A_a$ between two patches is $A_a = 1-d_a$. We use an appearance affinity threshold $\tau_a = 0.895$, determined offline, to separate similar and dissimilar pairs.
    
\begin{figure}[!b]
    \centering
    \subfigure[]{\label{fig:newmethod}\includegraphics[width=0.9\columnwidth]{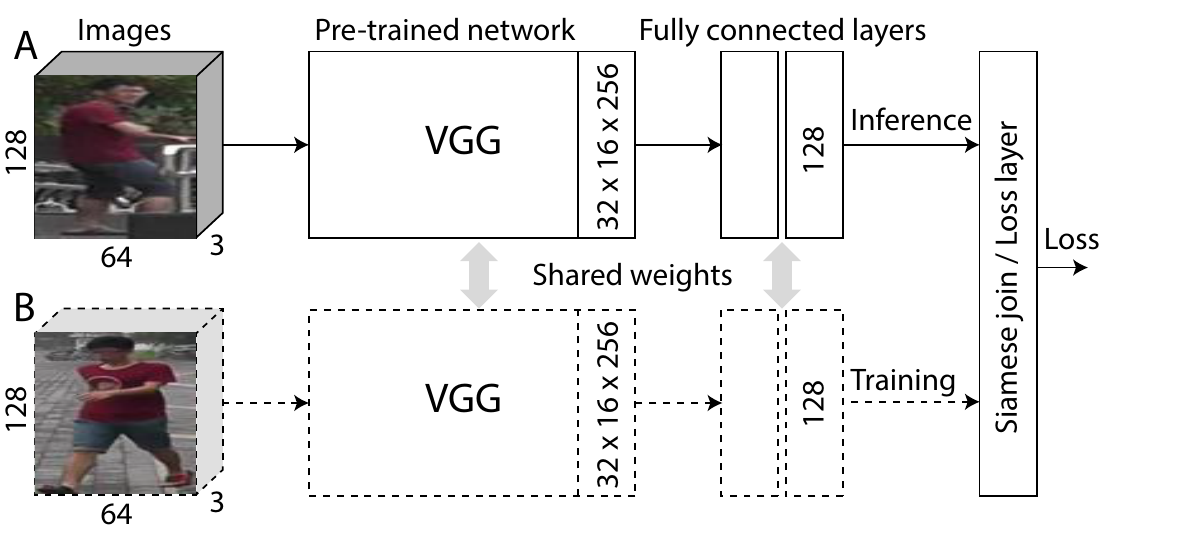}}
     \subfigure[]{ \label{fig:oldmethod}\includegraphics[width=0.9\columnwidth]{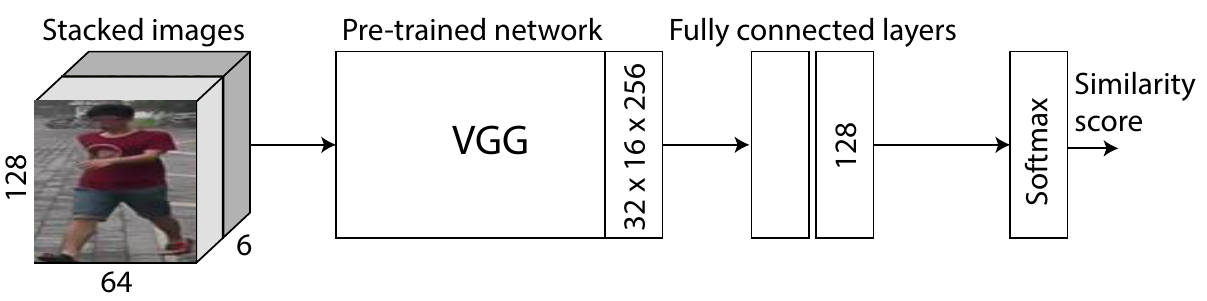}}
        \caption{(a) The proposed Siamese network that learns a similarity metric offline, with a margin contrastive loss, (b) An alternative Siamese network that takes two images as an input and outputs a similarity score.}
        
    \label{fig:twofigs}
    \end{figure}

    Two implementations of Siamese networks are shown in figure \ref{fig:twofigs}. Figure \ref{fig:newmethod} is the proposed implementation, using margin contrastive loss with a fixed margin of $0.2$. Figure \ref{fig:oldmethod} is an alternative implementation, using a learned softmax classifier to give a similarity score between the input images. In our approach, we compute the feature vector for a detection and store it with the state of the tracklet at the time of the detection, meaning that we don't have to store a gallery of images for each tracked object. We assume that the appearance metric computed from the detection will closely match the appearance metric of the true bounding box of the subject. This assumption appears to hold during testing, as bounding boxes are usually well regressed to the true bounding box of the detected object.

    \begin{table}[!b]

            \caption{Similarity Network Structure}
        \label{tab:nn_structure}
        \centering
        \begin{tabular}{lc}
        \toprule
             Layer & Output shape \\
             \midrule
             Input & $128 \times 64 \times 3$ \\
             VGG-16 & $32 \times 16 \times 256$\\
             Fully connected & $128$\\
             Fully connected & $128$ \\
             Batch normalisation & $128$\\
             $l_2$ normalisation & $128$ \\
             \bottomrule
        \end{tabular}
    \end{table}
    
    \subsection{Training}
    \label{sec:training}
    The deep similarity network was trained on the Market-1501 pedestrian re-identification dataset\cite{ZhengScalablePersonReidentification2015}, containing $\approx$ 32,000 annotated images of 1501 unique pedestrians in six camera views. Triplet loss has recently been used to good effect in training networks for pedestrian re-identification\cite{HermansDefenseTripletLoss2017}. Networks using triplet loss have been known to be difficult to train, due to a stagnating training loss. Batch-hard example mining has been shown to improve convergence when training with triplet loss\cite{HermansDefenseTripletLoss2017}. Our approach uses batch-hard sampling to train our network in a Siamese fashion using margin contrastive loss in a large batch. We sample 4 images each from 32 identities, compute their feature vectors in a forward pass and select the hardest pairings, maximising Euclidean distance between feature vectors for positive pairs and minimising distance for negative pairs, for each of the 128 images. 
    

    \begin{figure*}[t]
        \centering
        \includegraphics[width=1.0\textwidth]{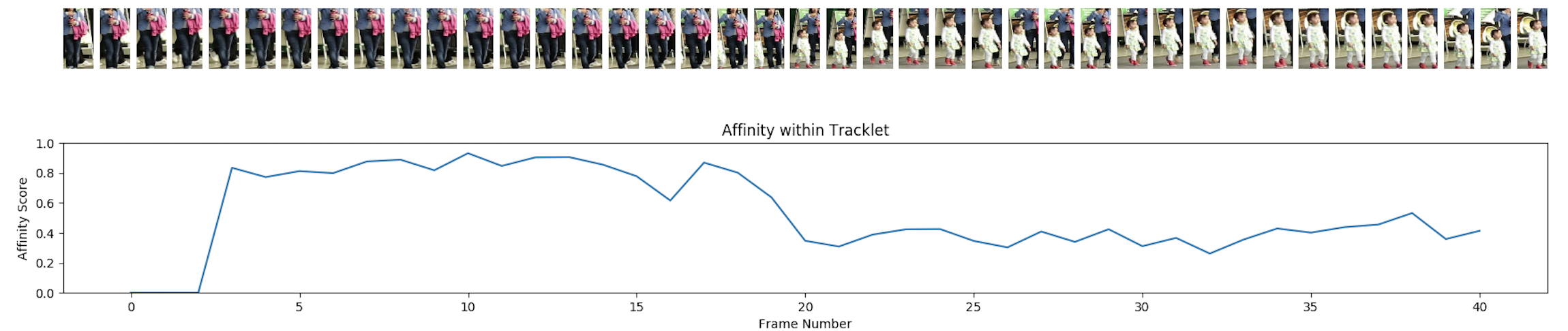}
        \caption{Using our deep similarity metric network, appearance information is stored while detections are strong. The proposed appearance affinity measure $A_a$ can detect and prevent tracklet drift and ID switches as shown in this example. The affinity between the track at $t=0$ and each detection drops below a level required for association as the tracker drifts to a new identity.}
        \label{fig:drifting_track}
    \end{figure*}
    
    \subsection{Detection Assignment} 
    \label{sec:assignment}
    Detections are combined across time to estimate the trajectory of a tracked object. This algorithm is shown in listing \ref{lst:assignment} and detailed below. The motion of small segments, tracklets, are estimated via a Kalman filter with a constant velocity constraint. Tracklet states are predicted at each time step, but are considered inactive after two predictions without being assigned a detection. A tracklet's state is predicted for another 90 steps for tracklet association, discussed in section \ref{sec:Tracklet_Association}. The association of new detections to the set of active tracklets is solved as a data association problem using the Hungarian algorithm\cite{Hungarianmethodassignment}. The Hungarian algorithm maximises the affinity between tracklets and assigned detections, provided in the affinity matrix $\tilde{A}$, and creates entries in the matching matrix $\tilde{M}$. The affinity used to assign candidate detections to tracklets is a combination of motion affinity, preferencing detections close to the predicted position of the tracklet, and appearance affinity which attempts to match the tracklet with a detection whose appearance is closest to stored appearance information.
    Motion affinity is implemented as the Intersection over Union (IoU)\cite{YuUnitBoxAdvancedObject2016} between a candidate detection $D_i^{t}$ and the predicted bounding box of the tracklet ${T_j^{t}}^{\prime}$, as shown in equation \ref{eq:motionAffinity}. Motion affinity is constrained to be strictly greater than a motion affinity threshold $\tau_m = 0.3$ for assignment.

    \begin{equation}
        A_m(T_j, D_i^t) = IoU({T_j^t}^{\prime}, D_i^t),\;\; \textit{IoU}(b_1, b_2) = \frac{b_1 \cap b_2}{b_1 \cup b_2}
        \label{eq:motionAffinity}
    \end{equation}

    Appearance affinity is computed as the mean affinity between a candidate detection's feature vector and the stored feature vectors for a tracklet, shown in equation \ref{eq:appearance_affinity} with $t_0$ denoting the first state of the tracklet. An example of appearance affinity degrading as a track drifts to an overlapping detection is shown in figure \ref{fig:drifting_track}. A subset of N past states of the tracklet is used for computational tractability, in practice $N \leq 20$.
    
    \begin{equation}
        A_a(T_j, D_i^{t}) = \frac{1}{N}\sum_{n = n_0}^{N} \> f(T_j^n, D_i^{t}),\;\; n \in \{t_0,t-1\}
        \label{eq:appearance_affinity}
    \end{equation}
    
    \begin{equation}
        f(T_j^n, D_i^{t}) = 1 - || f_1-f_2 ||
        \label{eq:euclidean}
    \end{equation}
    
    The total affinity, shown in equation \ref{eq:total_Affinity}, is a combination of appearance and motion affinity, balanced by the parameter $\lambda$, typically between $0.3$ and $0.7$. A value of $0$ may be used to ignore appearance entirely when computing assignments, potentially improving frame rate under some implementations.
    
    \begin{equation}
        A(T_j^t, D_i^t) = \lambda A_a(T_j, D_i^t) + (1-\lambda)A_m(T_j, D_i^t)
        \label{eq:total_Affinity}
    \end{equation}
    
    \begin{algorithm}[t]
    \begin{algorithmic}[1]
    \FOR{$T_j,D_i^t \in \mathcal{T},\mathcal{D}^t$}
        \IF{$A_m(T_j, D_i^t) > \tau_m \land A_a(T_j, D_i^t) > \tau_a$}
            \STATE $\tilde{A}_{j,i} \gets A(T_j, D_i^t)$ 
        \ELSE
            \STATE $\tilde{A}_{j,i} \gets 0$
        \ENDIF
    \ENDFOR
    \STATE $\tilde{M} \gets H(\tilde{A})$ \COMMENT{Hungarian Algorithm Matching}
    \FOR{$T_j,D_i^t \in \mathcal{T},\mathcal{D}^t$}
        \IF{$\tilde{M}_{j,i} = 1$}
            \STATE update state of tracklet $T_j$ with detection $D_i^t$
        \ENDIF
    \ENDFOR
    \end{algorithmic}
    \caption{Detection Assignment}
    \label{lst:assignment}
    \end{algorithm} 

    \subsection{Tracklet Confidence}
    A minimum length requirement $\tau_l = 6$ is imposed on tracklets for them to be considered positive. Tracks containing less than six states are considered negative and therefore are not reported. The mean confidence of detections assigned to a given tracklet is also used to filter out low confidence tracklets, with a minimum mean confidence of $\tau_c = 0.2$ used in practice. The average cost of assigning detections to a tracklet is used to estimate confidence in it being positive. A tracklet with a high mean assignment cost is likely to be varying in appearance or in motion and is considered negative. Tracklet association and boosting considers only positive tracks to avoid joining false positives with true positives.

    \subsection{Detection Boosting} \label{sec:boosting}
    In the case that in a given frame, there exists no detection which matches to a tracklet, but the tracked object is not occluded or out of frame, we wish to re-identify that person. Using the predicted location of the object as a prior, we perform dense sampling around the prediction and select the candidate bounding box which maximises appearance affinity and satisfies the appearance affinity constraint $\tau_a = 0.895$. This detection is added to the detection set and association is performed again, as shown in figure \ref{fig:OSTAB}. In order to prevent track drift, boosting is limited to no more than once per two frames per track. To stop partial detections from drifting to a true person via boosting and therefore adding false positives, Non Maximum Suppression (NMS) is performed on the detections with a NMS-IoU threshold of $0.5$.
    
    \begin{table*}[t]
    \caption{Results on the MOT16 test set, compared to all methods reported as online and over 15 FPS, a cutoff we suggest for real-time applications. We cite all non-anonymous methods. Best results in each category appear in bold.}
        \label{tab:MOT16}
        \centering
        \begin{tabular}{lccccc}

            \toprule
            Method & MOTA $\uparrow$ & FP $\downarrow$ & FN $\downarrow$ & IDs $\downarrow$ & FPS $\uparrow$ \\
            \midrule
            MOTDT\cite{long2018tracking}  & \textbf{47.6} & 9,253 & \textbf{85,431} & 792 & 20.6\\
            TestUnsup & 41.5 & 12,596 & 93,404 & 643 & 19.7 \\
            PMPTracker & 40.3 & 10,071 & 97,524 & 1,343 & 148.0 \\
            \textbf{SDMT (ours)} & 39.6 & 11,130 & 98,343 & \textbf{602} & 19.8\\
            RNN\_A\_P & 34.0 & 8,562 & 109,269 & 2,479 & 19.7 \\
            cppSORT\cite{murray2017real} & 31.5 & 3,048 & 120,278 & 1,587 & \textbf{687.1}\\
            DCOR  & 28.3 & \textbf{1,618} & 128,345 & 849 & 32.9\\
            \bottomrule
        \end{tabular}
    \end{table*}
    \subsection{Tracklet Association}
        \label{sec:Tracklet_Association}

    Targets may be tracked through occlusion by matching tracklets across time using their appearance. Our association algorithm is shown in listing \ref{lst:association} and described below. Due to uncertainties in camera and target motion, a much looser motion constraint is used to associate tracklets, requiring only a small overlap between the predicted bounding box of the older tracklet and the first bounding box of the newer track i.e. $IoU({T_j^t}^{\prime}, T_k^t) > 0$.
    
    \begin{equation}
        A_a(T_j, T_k) = \frac{1}{N} \frac{1}{M} \sum_{n = n_0}^{N} \sum_{m = m_0}^{M}\> f(T_j^n, T_k^m)
        \label{eq:tracklet_appearance}
    \end{equation}
    
    As tracking is done in the image plane, changes in camera motion may frequently violate the constant velocity constraint imposed by our Kalman filter based tracking. By building small tracklets with a stricter motion constraint and linking high confidence tracklets in to longer tracks with a looser motion constraint, intuitively our tracking may be robust to changes in camera motion. Tracklet association need not run at every time step, once every 20 time steps is sufficient to not impact performance, resulting in a higher refresh rate. 
    
    After tracklets have been merged, temporal gaps are filled by interpolation with a constant velocity, giving a reasonable estimate for the state of the object while it is occluded.

    \begin{algorithm}[t]
    \begin{algorithmic}[1]
    \FOR{$T_j \in \mathcal{T}$} \label{ln:beginning}
        \FOR{$T_k \in \mathcal{T} - T_j$}
            \IF{confidence constraints met on $T_j$ and $T_k$ temporal overlap exists}
                \IF{$IoU(T_j^t, {T_k^{t}}^\prime) > 0$}
                    \STATE $\mathcal{C} \gets \mathcal{C} \cup T_k$ \COMMENT{build set of candidate matches}
                \ENDIF
            \ENDIF
        \ENDFOR
        \STATE select best match\\ $T_o = \underset{l} {\mathrm{\text{arg\,max}}} \; A_a(T_j, T_l) \quad \forall \> T_l \in \mathcal{C} $
        \STATE merge tracklets \quad $T_j \gets T_j \cup T_o$
        \STATE \textbf{go to} \ref{ln:beginning}  
    \ENDFOR
    \end{algorithmic}
    \caption{Tracklet Association}
    \label{lst:association}
    \end{algorithm}

 \begin{table*}[th!]
    \caption{Ablative testing performed on the MOT16 training set, with and without appearance modelling for detection assignment, detection boosting, and tracklet association. Best results in each category appear in bold.}
        \label{tab:ablative}
        \centering
        \begin{tabular}{lccccc}
            \toprule
            Method & MOTA $\uparrow$ &  FP $\downarrow$ & FN $\downarrow$ & IDs $\downarrow$ & FPS $\uparrow$ \\
            \midrule
            SDMT ($\lambda = 0.5$) & \textbf{34.6} & 6,014 & 65,863 & 317 & 29.8\\
            SDMT ($\lambda = 0$) & 34.3 & 6,541 & 65,651 & \textbf{295} & 29.6\\
            SDMT ($\lambda = 1$) & 33.9 & 6,512 & 66,040 & 373 & 28.7\\
            SDMT (w/ boosting) & 34.2 & 6,743 & \textbf{65,533} & 334 & 25.9\\
            SDMT (w/o association) & 32.4 & \textbf{3,968} & 69,965 & 686 & 31.6\\
            SDMT (w/o appearance modelling) & 32.9 & 4,069 & 69,468 & 587 & \textbf{96.8}\\
            \bottomrule
        \end{tabular}
    \end{table*}
    
\section{Evaluation}
\label{sec:evaluation}
    
    
    The Siamese deep metric network was validated on a subset of Market-1501 dataset not used for training. The network achieved an area under the receiver operator characteristic curve of $0.98$ after $90,000$ training iterations, with an equal mix of positive and negative pairs and distractors sampled from the background. This validates the training of the deep similarity network.

    \subsection{CLEAR MOT Metrics}
    The CLEAR MOT\cite{StiefelhagenCLEAR2006Evaluation2006} metrics are used here to compare our performance to others, as well as compare the benefits of each of the uses of our deep appearance model. The specific metrics we use ($\uparrow$ denotes metrics in which a higher score is better, $\downarrow$ denotes metrics in which a lower score is better):
    \begin{itemize}
        \item MOTA $\uparrow$, combines FP, FN and IDs to give a single metric to summarise accuracy.
        \item FP $\downarrow$, is the number of false positive bounding boxes.
        \item FN $\downarrow$, is the number of false negative bounding boxes.
        \item IDs $\downarrow$, is the number of times tracked targets swap ID's.
        \item FPS $\uparrow$, is the update frequency, an important metric for real-time applications.
    \end{itemize}

    \subsection{MOT16 results}
    A selection of methods suitable for real-time applications was made for comparison. Online approaches that achieve an update rate of greater than 15\,Hz on the MOT16 test set using the public detections are shown in table \ref{tab:MOT16} compared to our approach. Among the comparable approaches, our method achieves a competitive tracking accuracy (MOTA) for a relatively simple method, but crucially the lowest number of ID switches, thanks to robust tracklet association and detection assignment.

    We performed repeated testing while enabling/disabling certain aspects of our algorithm, presented in table \ref{tab:ablative}. The best performing method from this ablative testing was used in testing presented in table \ref{tab:MOT16}. The best method did not include boosting and used a lambda value of $\lambda = 0.5$. Changing $\lambda$ to 0 or 1 reduced accuracy on the training set. Adding boosting to the optimal method reduced false negatives but significantly increased false positives. Removing tracklet association, or appearance modelling entirely significantly reduced tracking accuracy. The method without any appearance modelling removed the need to compute feature vectors for each detection, significantly increasing the update rate.

\section{Discussion}
\label{sec:discussion}
We found that using our deep appearance metric for detection assignment and tracklet association improved the overall performance of multiple object tracking. Only `detection boosting' was found to hurt the accuracy of our tracking on this dataset, despite reducing the number of false negatives as intended. This was likely due to the high recall rate of 43\% (with significant occlusion) but relatively low precision of the DPM v5 detections provided with the test sequences\cite{MilanMOT16BenchmarkMultiObject2016}. Boosting is most useful when there exists no detection for a given target, yet the target is not completely occluded or out of frame, however we suggest this case does not occur often in the MOT16 dataset. Tracklets built from false positive detections that contain some part of a true object, may be boosted, causing drift towards the true object. This may lead to the tracks being merged, explaining the increase in false positives.

We found that the addition of deep metric learning significantly reduced the number of ID switches. The ablative study suggests that the largest change in ID switches was due to tracklet association. This reflects the major benefit of a tracking formulation with a strong appearance model, estimating the position of objects while they are occluded. e.g. A pedestrian walking behind a bus.

\section{Conclusions}
\label{sec:conclusions}
We presented three uses of deep appearance metric learning for improving multiple object tracking, and demonstrated how two of these uses significantly improved tracking accuracy on the MOT16 dataset. Our method achieved competitive results for online methods suitable for real-time applications, with the lowest number of ID switches. Our ablative testing may be used to inform further use of deep appearance metrics in multiple object tracking.

\balance

\bibliographystyle{named}
\bibliography{ostab}
\end{document}